\documentclass[conference,a4paper]{IEEEtran}
\IEEEoverridecommandlockouts
\usepackage{cite}
\usepackage{amsmath,amssymb,amsfonts}
\usepackage{algorithmic}
\usepackage{graphicx}
\usepackage{textcomp}
\usepackage[ruled,vlined]{algorithm2e}
\usepackage{lipsum}
\usepackage{balance}
\usepackage{multicol, blindtext}
\usepackage{xcolor}
\usepackage[utf8]{inputenc}

\usepackage{tikz}
\usepackage[utf8]{inputenc}
\usepackage{textcomp}
\usepackage{pgfplots,capt-of}
\usepackage{pgfplots}
\begin{document}

\title{\LARGE Exploiting Unlabeled Data in Smart Cities using Federated Edge Learning}
\author{\IEEEauthorblockN{Abdullatif Albaseer, Bekir Sait Ciftler, Mohamed Abdallah, and Ala Al-Fuqaha}
\IEEEauthorblockA{Division of Information and Computing Technology, College of Science and Engineering,
\\Hamad Bin Khalifa University, Doha, Qatar \\
amalbaseer@mail.hbku.edu.qa,
\{bciftler, moabdallah, aalfuqaha\}@hbku.edu.qa}
}
\maketitle
\vspace{-8mm}
\begin{abstract}
Privacy concerns are considered one of the main challenges in smart cities as sharing sensitive data induces threatening problems in people's lives.
Federated learning has emerged as an effective technique to avoid privacy infringement as well as increase the utilization of the data.
However, there is a scarcity in the amount of labeled data and an abundance of unlabeled data collected in smart cities; hence there is a necessity to utilize semi-supervised learning.
In this paper, we present the primary design aspects for enabling federated learning at the edge networks taking into account the problem of unlabeled data.
We propose a semi-supervised federated edge learning method called FedSem that exploits unlabeled data in real-time.
FedSem algorithm is divided into two phases. The first phase trains a global model using only the labeled data.
In the second phase, Fedsem injects unlabeled data into the learning process using the pseudo labeling technique and the model developed in the first phase to improve the learning performance.
We carried out several experiments using the traffic signs dataset as a case study. Our results show that FedSem can achieve accuracy by up to 8\% by utilizing the unlabeled data in the learning process.
\end{abstract}


\section{Introduction}
Smart cities provide reliable and robust solutions to crucial problems related to traffic, healthcare, education, security, etc~\cite{amma2018privacy,puiu2016citypulse}.
Smart cities embody a massive smart Internet of things (IoT) devices in various applications.
The compelling capabilities of the sensors included in these devices generate an unprecedented volume of data~\cite{brisimi2016sensing}.
Learning from this data reinforces the performance of applications and enables the discovery of the knowledge to compose intelligent decisions~\cite{amma2018privacy}.
However, a large chunk of this data is sensitive because it is generated by users, and their privacy is a premier parameter that must be fulfilled in the design of smart cities' infrastructure to evade privacy infringement\cite{khan2019federated}.
In addition, sending massive data to a centralized location is resource starvation resulting in network congestion since many users endeavor to make use of the same resource~\cite{mcmahan2016communication}. To this end, there is a need for a distributed learning paradigm that mitigates network bottlenecks and enables IoT devices to build a collaborative shared model that discovers the necessary information embedded in the data without compromising their privacy.

Federated learning (FL) has emerged as an attractive solution to meet the aforementioned requirements\cite{tran2019federated}.
FL enables users to share their acquaintance without privacy violation, whereas the data is stored locally\cite{mcmahan2016communication}.
Users only share their local model gradients regularly with the orchestrating server, which organizes the training and collects the contributions of all participants~\cite{smith2018cocoa}.
The server builds the global model by averaging all gradients across the network~\cite{Sahu2018FederatedOF}. Then, the coordinating server broadcasts the new updated model to all clients ~\cite{lim2019federated}. Each client uploads its local model to the server and then downloads the global model to do on-device inference using a cloud-distributed model. The server orchestrates this process until learning is stopped~\cite{smith2017federated}.

Moreover, to allow rapid access to the enormous distributed data for fast model training, federated learning algorithms have been pushed towards the network edge.
This has led to the emergence of a new paradigm of FL called federated edge learning as a cutting-edge decentralized technique that enables edge devices to train the model using real-time data collaboratively~\cite{konevcny2016federated}.

Recently, the works in~\cite{mcmahan2016communication,smith2017federated, Sahu2018FederatedOF,lin2018don,lim2019federated } studied system-level and statistical challenges related to FL deployment as detailed in Section~\ref{sect:RelatedWorks}.
However, none of the existing works ~\cite{mcmahan2016communication,smith2017federated, Sahu2018FederatedOF,lin2018don,lim2019federated } considers the unlabeled data, and they only assumed that the data is completely labeled which does not reflect the realistic nature of the applications.
In reality, there is a scarcity in the amount of labeled data and an abundance of unlabeled data collected in smart cities.

To this end, targeting federated edge learning, the main contribution of this work is a novel semi-supervised federated learning scheme that exploits the unlabeled data at edge networks.
In our simulation, we use German Traffic Sign Dataset (GTSDB) which contains large images of real-world traffic signs to evaluate the proposed approach.
We can summarize our key contributions as follows:
\begin{itemize}
\item We propose a novel semi-supervised FL approach called FedSem, FedSem can handle the problem of unlabeled data in smart cities while preserving privacy and increasing data utilization.
\item We utilize the GTSDB dataset to evaluate our proposed method under various settings of unlabeled data ratios.
\item We consider the performance of FL under various heterogeneity settings for the unlabeled data.
\end{itemize}

To the best of our knowledge, this is the first work in FL that takes into account semi-supervised learning, which exploits unlabeled data generated in the edge networks.

The rest of this paper is structured as follows.
We review the state of art of similar works in Section~\ref{sect:RelatedWorks}.
Then, we introduce the system model and our proposed approach in Section~\ref{sect:SystemModel}.
Details descriptions of the used datasets, performance metrics, experimental setup, and results are provided in Section~\ref{sect:NumericalResults}.
Finally, we conclude our work with remarks in Section ~\ref{conclustion}.

\section{Related Work}
\label{sect:RelatedWorks}
To process large amounts of data, with the evolution of cloud computing techniques, the majority of the works have been devoted to study large-scale distributed learning, especially in the data center setting ~\cite{boyd2011distributed, dean2012large, dekel2012optimal, shamir2014communication}.
However, pushing the data directly to the server violates the privacy of users for critical applications.
Recently, FL has emerged as an effective solution to preserve privacy and share the knowledge between users due to the rapid growth of computing agents (i.e., smartphones, wearables, and internet-of-things devices)~\cite{smith2017federated}.
In this approach, it is directly learning the models over the network rather than transmitting the data to the cloud~\cite{smith2017federated}.
This technique inspired researchers to pay attention to challenges with heterogeneity, privacy, computation constraints, and communication cost.

In order to evaluate the proposed methods in FL, researchers have taken into account the following properties ~\cite{mcmahan2016communication}:
\begin{itemize}
\item Non-IID: training the data in each client depends on its usage. Consequently, any specific user's local data could not represent the population distribution~\cite{smith2017federated}
\item Unbalanced data: the local training data (e.g., Sign Images) is varied depending on the usage of the service, which depends on the user behaviour~\cite{Sahu2018FederatedOF}.
\item Communication boundaries: Some devices(e.g., smartphones, vehicles) typically are not available all the time or may have slower connections. Also, different users may use different network technologies(i.e. 4G and 5G)~\cite{smith2017federated}.
\end{itemize}

Focusing on optimization algorithms for FL, many methods ~\cite{boyd2011distributed, lin2018don} have been designed to tackle the statistical and system challenges.
These methods showed outstanding improvements compared to conventional approaches such as ADMM methods~\cite{boyd2011distributed} and mini-batch~\cite{dekel2012optimal} algorithm.
These methods allow for local updates in the edge devices by only activating a subset of them to participate in forming a global model~\cite{smith2017federated, lin2018don}.
In addition, with the aim of convergence, the authors in ~\cite{smith2018cocoa} proposed a heuristic method called multitask learning to average the local updates received from a set of devices and then broadcast the global model accordingly.
The authors proposed to collect raw data in a certain period to improve the model.
However, the data is private, and this will violate the principle of FL, which mainly aims to preserve the user's privacy.

Recently, heuristic methods have been proposed to address statistical data heterogeneity in FL ~\cite{lin2018don, mcmahan2016communication}.
For example, Federated Averaging (FedAvg) is a heuristic algorithm based on averaging local Stochastic Gradient Descent (SGD) updates in the primal.
In ~\cite{mcmahan2016communication}, the authors showed the FedAvg method of providing outstanding empirical performance.
However, FedAvg is challenging to analyze due to its local updates in regular periods, and only a subset of devices participates at each round with heterogeneous data in non-identical fashion.
To tackle this issue, the authors in ~\cite{smith2017federated, smith2018cocoa} proposed approaches to periodically send the local data produced by the edge devices to edge-server and then, share the global model to all edge devices.
However, these methods are unrealistic because the bandwidth and energy are quickly consumed due to periodic data transmissions, and user privacy is violated.
On the other side, sharing the edge-device data between all members requires sufficient network resources and powerful computing capabilities to manipulate massive datasets. Furthermore, a new paradigm of FL called federated edge learning has emerged as a cutting-edge decentralized technique which enables edge devices to train the model using real-time data~\cite{lim2019federated} collaboratively.

In summary, the majority of researchers have studied the statistical challenges of FL.
However, to the best of our knowledge, there are no approaches to handle the problem of using unlabeled data collected in smart cities using edge FL.

\section{System Model}
\label{sect:SystemModel}
\begin{figure}[t]
\centering
\includegraphics[width=0.95\linewidth, height=6cm]{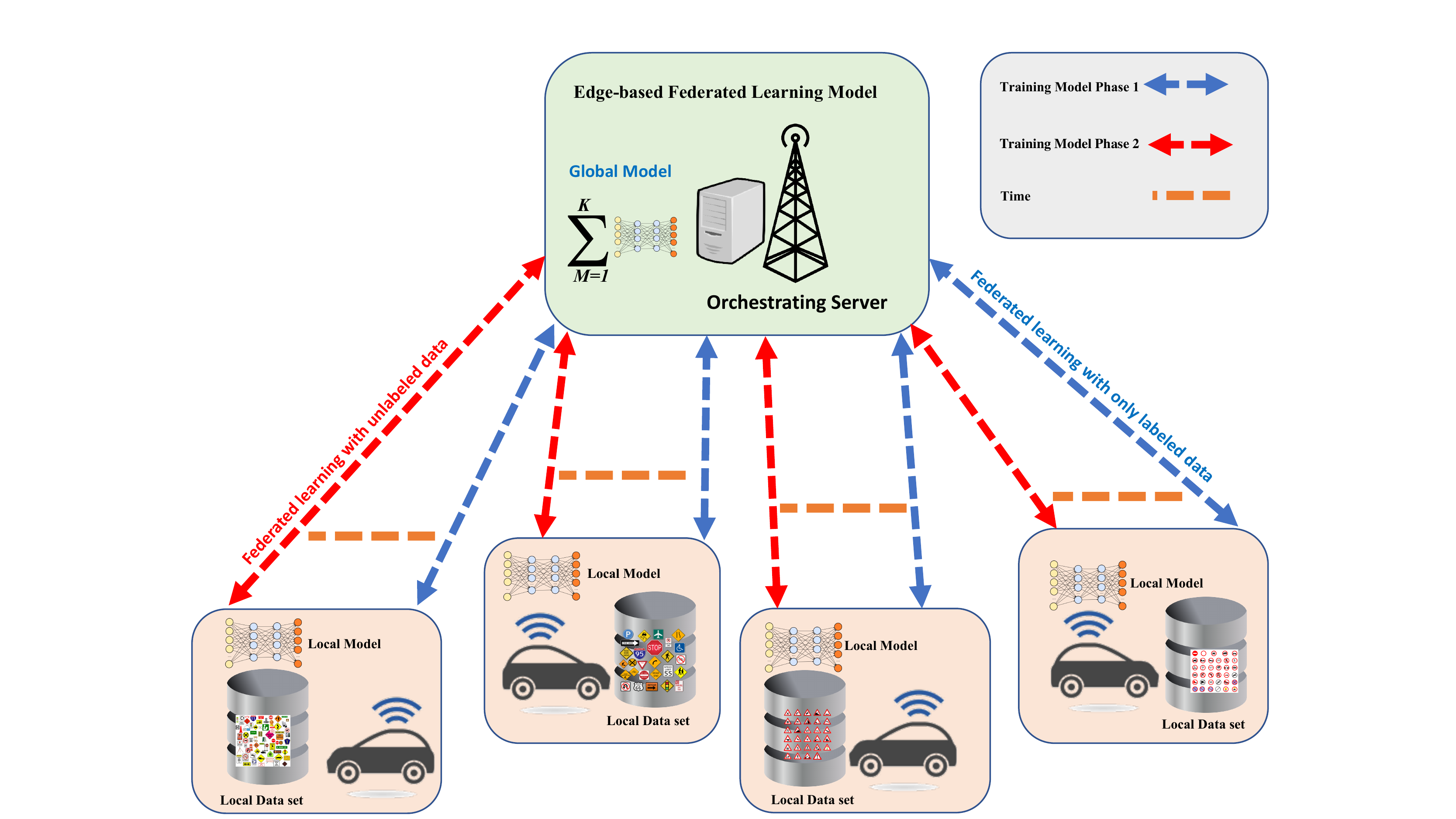}
\caption{System model under Federated learning settings}
\label{fig1}
\vspace{-5mm}
\end{figure}

In this work, we consider a swarm of smart vehicles learning traffic signs at edge networks in smart cities.
This system includes a set of autonomous vehicles $K$ passing through different roads, as depicted in Fig~\ref{fig1}.
A subset of these vehicles $S$ occasionally is active.
The coordination between these nodes is performed by an orchestrating edge server that controls the learning process.
Each vehicle trains the local model based on its own data (traffic sign images) locally and sends only gradients to the server.
Then, the server applies federated averaging to create a global model using \eqref{eq1}.
In general, the orchestrating server coordinates to select different number of participants in each round trying to capture all existing labels across vehicles.
\begin{equation}
\omega^{t+1}= \frac{1}{K}\sum_{k\in S_t} \omega^{t+1}_k \label{eq1}
\end{equation}
where $w$ is the weights of the global model, $K$ is the total number of vehicles in the network, and $S_t$ is the subset of vehicles selected to train the global model for one round.

To train the local model across vehicles, the loss function is defined as follows.
Consider ${\cal D}_k$ denotes the local dataset collected at the $k$-th edge vehicle. The loss function of the model $\omega$ on ${\cal D}_k$ is given by
\begin{align}\label{eq:local_loss} (\text{Local loss function}) \qquad
F_k(\omega) = \frac{1}{|{\cal D}_k|} \sum_{(x_j, y_j) \in {\cal D}_k} f(\omega, x_j, y_j)
\end{align}
Then, the loss function of all vehicle is expressed as follows.
\begin{align} (\text{Global loss function}) \qquad
F(\omega) = \frac{1}{K} \sum_{k\in S_t} F_k(\omega).
\end{align}

All vehicles aim to minimize the global loss function $F(\omega)$, namely,
\begin{align}\label{eq:learning_prob}
\omega^* = \arg \min F(\omega).
\end{align}

\subsubsection{Federated Averaging}(FedAvg) ~\cite{mcmahan2016communication}, In FedAvg, the stochastic gradient descent $(SGD)$ is used as a local solver and each vehicle $k$ has a local surrogate to approximate the global objective function.
The local solver hyper-parameters(i.e., learning rate and local epochs) are assumed to be homogeneous among all vehicles in all rounds $R$. At each round $r$, only a subset of $K$ participants is selected to update the global model.
The $SGD$ is run locally for a specified number of epochs $E$ and the learning rate $\eta$.
A central server repeats these steps until convergence.
The steps of this approach are listed in Algorithm ~\ref{algavg}.
\begin{algorithm}[t]
\SetAlgoLined
\label{algavg}
\caption{FedAvg~\cite{mcmahan2016communication}}
\KwIn{$R, K, \eta , \omega^0, S, E$}
\For{$r= 1$ to $R$}{
1- Server coordinates to choose subset $S$ of K randomly\\
2- Server broadcasts $\omega^t$ to $S$\\
3- Vehicle $k_i$ run a local solver for $E$ epochs to update $\omega^t$ with step size $\eta$ to get $\omega^{t+1}$ \\
4- The selected vehicle $k_i$ send its updated model $\omega^{t+1}$ back to the server\\
5- Server receive all updates from $S$ and average $\omega's$ as $\omega^{t+1}= \frac{1}{K}\sum_{k\in S_t} \omega^{t+1}_k$
}
\end{algorithm}

\section{Federated Semi-Supervised Learning}

In this section, we explain the semi-supervised learning in general then, we narrow this definition to FL settings.
Semi-supervised learning is an approach where unlabeled data is used to gain more understanding of the general structure~\cite{li2018pseudo}.
The labeled data plays an important role to classify unlabeled data based on the initial training model.
However, semi-supervised learning under FL is challenging as a single unit can't capture all labels independently as the data points are disseminated across the network.
To this end, we propose a novel scheme that addresses this issue under FL settings as described in section~\ref{fedsemd}
\begin{figure}[t]
\centering
\includegraphics[width=0.75\linewidth]{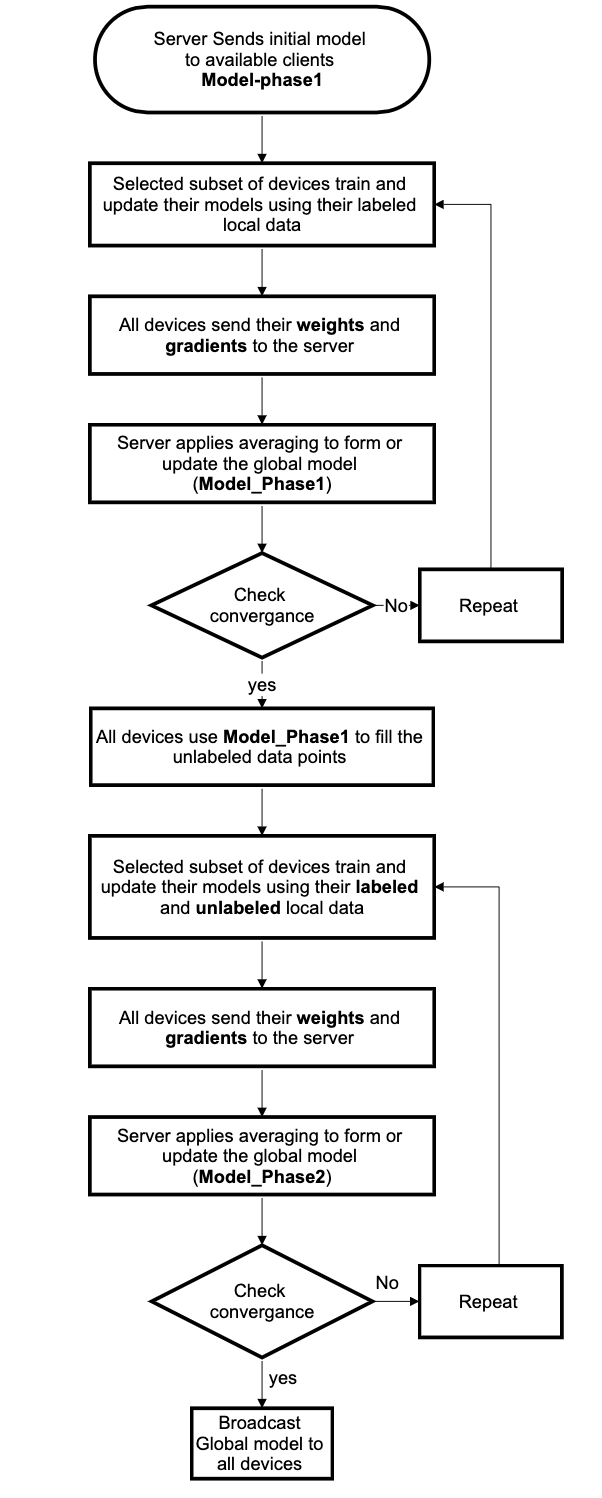}
\caption{FedSem Proposed Algorithm where the \emph{phase1} designs the global model using only labeled data and \emph{Phase2} injects unlabeled data into learning process}
\label{fig:FedSem}
\vspace{-8mm}
\end{figure}

\subsection{FedSem}
\label{fedsemd}
FedSem aims to leverage the semi-supervised learning process in edge FL.
We take advantage of using \emph{pseudo-labeling technique} to utilize unlabeled data in all vehicles in the network.
In the beginning, the server sends initial model \textbf{\emph{Model-Phase1}} with random gradients to available clients, which in turn will start training or updating their model using only their labeled data in order to collaboratively design the initial global model Model-Phase1.
This phase aims to capture all labels from different vehicles to ensure that the global model can predict most existing labels.
Then, the server control this phase until \textbf{\emph{Model-Phase1}} converges to enable all vehicles utilizing unlabeled data in the second phase.
In phase two, the resulting \textbf{\emph{Model-Phase1}} is used to fill the unlabeled data points.
As a result, all data is completely labeled and each vehicle uses traditional supervised learning after injecting unlabeled data into the training process so as to increase the robustness of the global model \textbf{\emph{Model-Phase2}}.
Fig.~\ref{fig:FedSem} is a flowchart illustrating our proposed scheme "FedSem".

For simplicity, Algorithm~\ref{fedsem_1} and Algorithm~\ref{fedsem_2} explain the steps in each phase.
\begin{algorithm}
\SetAlgoLined
\label{fedsem_1}
\caption{Federated Algorithm for Semi-supervised learning Phase-1}
\KwIn{Total Participant vehicles $ K$, Subset Participant in each round $S$, Learning rate $\eta $ , initial gradients $\omega^0$, Number of epochs in local vehicle $E$, Number of rounds $R$ }
\For{$i= 1$ to $R$}{
- Server coordinates to choose subset $S$ of K randomly\\
- Server broadcasts $\omega^t$ to $S$\\
\For{$j= 1$ to $S$}{
- Vehicle $k_j$ trains its model using only the fully labeled data points for $E$ epochs to update $\omega^t$ with step size $\eta$ to get $\omega^{t+1}$ \\
- The selected vehicle $k_j$ send its updated model $\omega^{t+1}$ back to the server\\
}
- Server receives all updates from $S$ and applies FedAvg $\omega's$ as $\omega^{t+1}= \frac{1}{K}\sum_{k\in S_t} \omega^{t+1}_k$\\
- \If{If the global model converge}{
Save the model "Model-Phase1"\;
Break\;
}
}
\end{algorithm}
\begin{algorithm}
\SetAlgoLined
\label{fedsem_2}
\caption{Federated Algorithm for Semi-supervised learning Phase-2}
\KwIn{Total Participant vehicles $ K$, Subset Participant in each round $S$, Learning rate $\eta $ , initial gradients $\omega^0$, Number of epochs in local vehicle $E$, Number of rounds $R$ }
- All $K$ vehicles use "Model-Phase1" to fill unlabeled data points.
\For{$i= 1$ to $R$}{
- Server coordinates to choose subset $S$ of K randomly\\
- Server broadcasts $\omega^t$ to $S$\\
\For{$j= 1$ to $S$}{
- Vehicle $k_j$ train its model utilizing labeled and unlabeled data for $E$ epochs to update $\omega^t$ with step size $\eta$ to get $\omega^{t+1}$ \\
- The selected vehicles $S$ send their updateds model $\omega^{t+1}$ back to the server\\
}
- Server receives all updates and applies FedAvg: $\omega's$ as $\omega^{t+1}= \frac{1}{K}\sum_{k\in S_t} \omega^{t+1}_k$\\
-\If{If the global model converge}{
Save the model "Model-Phase2"\;
Break\;
}
}
\end{algorithm}

\section{Numerical Results}
\label{sect:NumericalResults}

In this section, the performance of Fedsem method is benchmarked to the learning without injecting unlabeled data under different scenarios.
First, we explain the used dataset, how is distributed across vehicles and its structure, and then we present the used model and classifier.
\subsection{Dataset, Performance Metrics, and Experimental Setup}
\label{dataset}
The German Traffic Sign Dataset (GTSDB) has been widely used in similar research for only centralized supervised learning~\cite{sermanet2011traffic,staravoitau2018traffic}.
We follow the procedure done in ~\cite{sermanet2011traffic},~\cite{staravoitau2018traffic} for splitting the dataset.
The data is split into $39209$ 32×32 px color images for training and $12630$ images for testing.
Each image represents one of 43 distinct classes of traffic signs.
Each image is a 32×32×3 array of pixel intensities, represented as [0, 255] integer values in RGB color space.
The class of each image is converted to a one-hot encoding scheme.
We used a deep neural network classifier as a model following the work done in~\cite{sermanet2011traffic}.
For federated settings, we split the data between $1000$ vehicles, and in each round, only $30$ vehicles are selected randomly to train and update the model.
To assure data heterogeneity, the data is distributed in none-IID fashion.
For local image recognition, each vehicle uses convolutional neural networks as in ~\cite{li2018real}.
Table ~\ref{Tab:datasettings} illustrates GTSDB dataset and how it is split across vehicles.
\begin{table}[t]
\centering
\caption{Statistics of the used datasets in federated settings }
\begin{tabular}{|p{1.5cm}|p{1.5cm}|p{1.5cm}|p{1.5cm}|}
\cline{1-4}
\textbf{ Dataset} & \textbf{Total number of vehicles} & \textbf{Total number of samples} &\textbf{Number of classes} \\ \hline
\cline{1-4}
GTSDB & 1000 &39,209 for training, and 12,630 images for testing & 43 \\ \hline
\end{tabular}
\label{Tab:datasettings}
\end{table}
In this work, we use different percentages of the labeled data to show to which extent FedSem can help to enhance the learning performance. We use testing accuracy, testing loss and gained accuracy as performance metrics to evaluate FedSem.

We carried out all experiments using the TensorFlow library ~\cite{abadi2016tensorflow}.
$Adam$ optimizer is employed as a local solver.
The sampling scheme is implemented as in algorithms 1 and 2, which is uniform among vehicles.
The update is performed based on the weights to the local data points, as proposed in~\cite{mcmahan2016communication}.
For the model training parameters, we adopted the parameters similar to work in ~\cite{staravoitau2018traffic}.
However, we reduce the batch size to fit vehicle computation capabilities. We set homogeneous learning rate $\eta$ and the number of local epochs $E$ across vehicles.
The model has four layers comprising three convolutional layers for feature extraction and one fully connected layer as a classifier~\cite{staravoitau2018traffic}.

For our simulations, we consider that different participants are selected in each round to allow for more updates.
We have split the data on each vehicle into a training set $(80\%)$ and testing set $(20\%)$.
All matrices are reported using the global model outputs. The other parameters used in simulations are summarized in Table ~\ref{tab:setuppar}.
\begin{table}[t]
\centering
\caption{Experimental Setup parameters }
\begin{tabular}{|p{3.5cm}|p{2.5cm}|}
\cline{1-2}
\textbf{ Parameter} & \textbf{Value} \\ \cline{1-2}
Library & TensorFlow GPU \\ \cline{1-2}
Number of local epochs $E$ & 20, 40 and 100 \\ \cline{1-2}
Learning rate $\eta$ & 0.0001 \\ \cline{1-2}
Batch Size & 32 \\ \cline{1-2}
Number of rounds & 30, 50, and 100 \\ \cline{1-2}
Clients per round & 10 and 20 \\ \cline{1-2}
Evaluation Period & every round \\ \cline{1-2}
\end{tabular}
\label{tab:setuppar}
\end{table}
We initially carried our simulations using only $10\%$ of labeled data.
Then we repeat the same experiments with the same settings, but we use $30\%$ and $50\%$ of labeled data, respectively.
We set all experiments to start injecting unlabeled data at round $R/2$ where $R$ is the total number of rounds.

\subsection{Results}

In Table ~\ref{table:results}, we show the gained accuracy in both phases using different percentages of labeled data and a different number of epochs.
We compute the gained accuracy after injecting unlabeled data using (~\ref{equ:gain}).
\begin{equation}
\label{equ:gain}
Gain= \frac{Accu_{Phase2}-Accu_{Phase1}}{Accu_{Phase2}}
\end{equation}
where $Accu_{Phase1}$ is the achieved accuracy using only \textbf{\emph{Model-Phase1}} and $Accu_{Phase2}$ is the achieved accuracy after injecting unlabeled data into learning process \textbf{\emph{Model-Phase2}}.
We can notice that regardless of the percentage of the labeled data used in phase one, training the model using unlabeled data helps to increase the accuracy.
Also, we can observe that increasing the number of epochs across vehicles results to improve the testing accuracy so tuning the number of epochs to an optimal value is crucial to decrease the number of rounds that is needed to converge.
\begin{table}[t]
\centering
\caption{Experimental Results}
\begin{tabular}{|p{1cm}|p{1cm}|p{1cm}|p{1cm}|p{1cm}|p{1cm}|}
\cline{1-6}
\textbf{\% of Labeled data} & \textbf{\# of rounds} & \textbf{\# of Epochs $E$} & \textbf{labeled data Accuracy } & \textbf{All data points Accuracy } & \textbf{Gain} \\ \hline
\cline{1-6}
20 & 50 & 20 & 73\% & 78\% & 7\%\\ \hline
\cline{1-6}
20 & 50 & 40 & 80.02\% & 83.97\% & 5\%\\ \hline
\cline{1-6}
20 & 30 & 20 & 69.57\% & 73.55\% & 6\%\\ \hline
\cline{1-6}
30 & 50 & 20 & 79\% & 83\% & 5\%\\ \hline
\cline{1-6}
30 & 50 & 40 & 81\% & 84.7\% & 4.4\%\\ \hline
\cline{1-6}
50 & 50 & 40 & 79\% & 84.05\% & 6\%\\ \hline

\end{tabular}
\label{table:results}
\end{table}

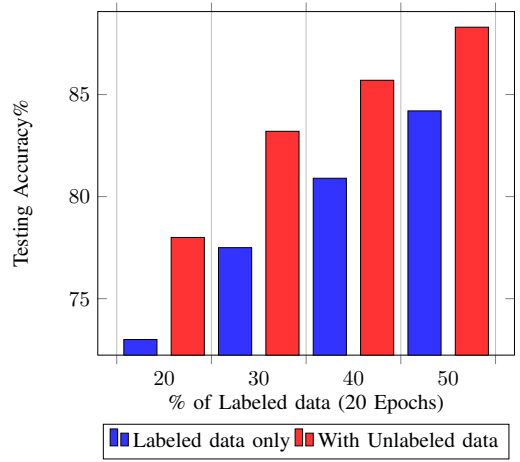
\begin{figure}[t]
\centering
\begin{tikzpicture}[scale=0.80]
\begin{axis}[
x tick label style={
/pgf/number format/1000 sep=},
xlabel=\% of Labeled data (20 Epochs),
ylabel=Testing Accuracy\%,
enlargelimits=0.05,
legend style={at={(0.5,-0.2)},
anchor=north,legend columns=-1},
ybar interval=0.7,
]
\addplot[fill=blue!80!white,]
coordinates {(20,73)
(30,77.5) (40,80.9)(50,84.2)(60,87)};
\addplot[fill=red!80!white,]
coordinates {(20,78)
(30,83.2) (40,85.7)(50,88.3)(60,88.1)}; 
\legend{Labeled data only, With Unlabeled data}
\end{axis}
\end{tikzpicture}
\caption{The percentage of the labeled data points vs. the testing accuracy when the number of Epochs is 20.}
\label{resultspre}
\end{figure}

\begin{figure}[t]
\centering
\includegraphics[width=0.8\linewidth]{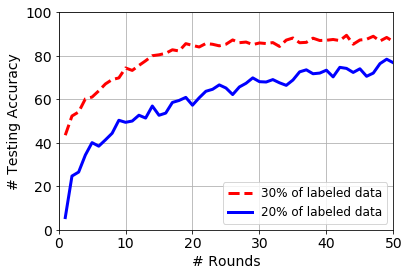}
\caption{The testing accuracy using FedSem vs 20\% and 30\% labeled data.}
\label{fig30p1}
\end{figure}
\begin{figure}[t]
\centering
\includegraphics[width=0.8\linewidth]{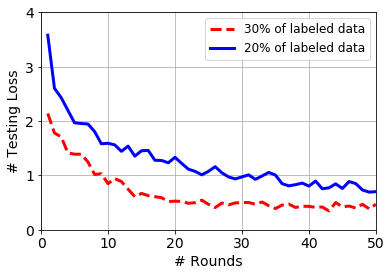}
\caption{The testing loss using FedSem vs. 20\% and 30\% labeled data.}
\label{fig30p2}
\end{figure}
Fig.~\ref{resultspre} shows the gained accuracy for different percentages of labeled data either without using unlabeled data or after injecting unlabeled data.
We can see that unlabeled data helps to increase the testing accuracy of all considered labeled data percentages.
This is because, at each given learning step, the proposed FedSem includes different features that belong to the same class, which in turn increases the marginal probability.

Fig.~\ref{fig30p1} and Fig.~\ref{fig30p2} show the obtained accuracy when the percentage of the labeled data is 30\% and 20\% during the learning process.
We can observe that utilizing unlabeled data enhances the accuracy and leverages the stability of the global model, as depicted in Fig.~\ref{fig30p1} and Fig.~\ref{fig30p2}.
In summary, injecting unlabeled data into training increases the accuracy even if the ratio of labeled data is small.

\section{CONCLUSIONS}
In this work, we propose a federated semi-supervised learning technique to utilize the unlabeled data in smart cities. The proposed scheme exploits the unlabeled data as well as evades privacy infringement.
The proposed approach divides the learning into two phases to assure capturing the information encapsulated within unlabeled data.
The global model resulting from \emph{Phase-1} is used to label the unlabeled data.
We have carried out several experiments using different percentages of labeled data to show how the FedSem can enhance the learning performance by utilizing unlabeled data even if the ratio of labeled data is small.
FedSem improves the accuracy of up to 8\% compared to using only the labeled data.
Overall, utilizing unlabeled data in FL increases accuracy. 
\label{conclustion}
\balance
\bibliographystyle{IEEEtranTIE}
\bibliography{ref}
\end{document}